\documentclass[letterpaper]{article} 
\usepackage[submission]{aaai2026}  
\usepackage{times}  
\usepackage{helvet}  
\usepackage{courier}  
\usepackage[hyphens]{url}  
\usepackage{graphicx} 
\usepackage{natbib}  
\usepackage{caption} 
\usepackage{multirow}
\usepackage[ruled]{algorithm2e}
\usepackage{amsmath}
\usepackage{amsfonts}
\usepackage{xcolor}

\usepackage{algorithmic}

\usepackage{newfloat}
\usepackage{listings}
\title{HOSIG: Full-Body Human-Object-Scene Interaction Generation with Hierarchical Scene Perception}
\author{
    Wei Yao\textsuperscript{\rm 1},
    Yunlian Sun\textsuperscript{\rm 1}\equalcontrib,
    Hongwen Zhang\textsuperscript{\rm 2},
    Yebin Liu\textsuperscript{\rm 3},
    Jinhui Tang\textsuperscript{\rm 4}\equalcontrib
}
\affiliations{
    \textsuperscript{\rm 1}Nanjing University of Science and Technology\\
    \textsuperscript{\rm 2}Beijing Normal University\\
    \textsuperscript{\rm 3}Tsinghua University\\
    \textsuperscript{\rm 4}Nanjing Forestry University\\
    \{wei.yao, yunlian.sun, jinhuitang\}@njust.edu.cn, zhanghongwen@bnu.edu.cn, liuyebin@mail.tsinghua.edu.cn

}
\makeatletter
\def\showauthors@on{T}  
\makeatother

\usepackage{bibentry}

\begin{document}

\twocolumn[{%
\renewcommand\twocolumn[1][]{#1}%
\maketitle

\begin{center}
    \centering
    \includegraphics[width=0.99\textwidth]{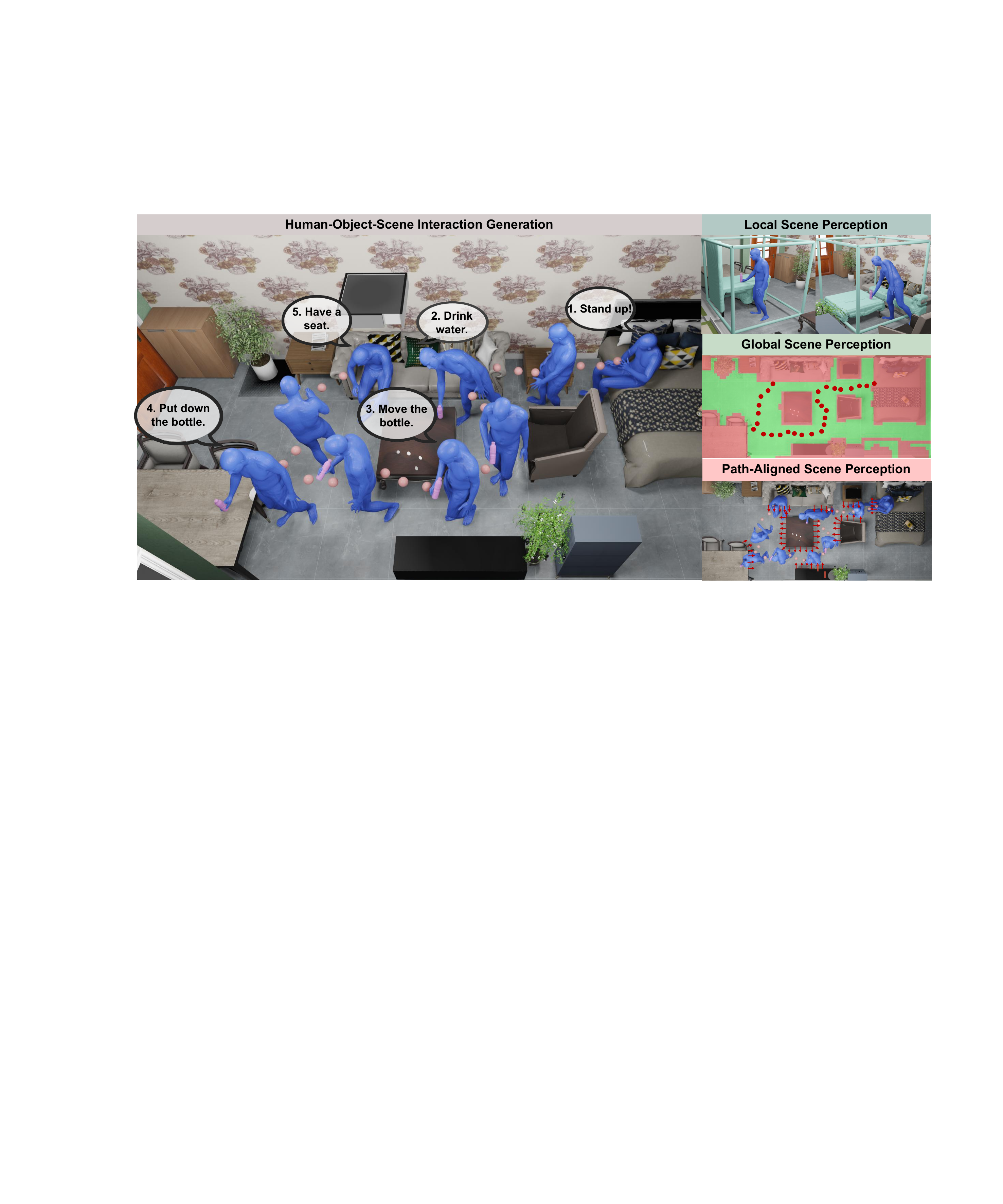}
    \captionof{figure}{\textbf{Human-Object-Scene Interaction Generation}. Our proposed HOSIG can generate high-fidelity full-body human motions. HOSIG can not only generate \textbf{interactions with static scenes}, but also generate object manipulation motions with \textbf{fine hand-object contact}. Moreover, relying on iterative generation and autonomous navigation, HOISG can generate \textbf{long-term} motions in complex indoor scenes.}
   \label{fig:teaser} 
\end{center}%
}]

\begin{abstract}
Generating high-fidelity full-body human interactions with dynamic objects and static scenes remains a critical challenge in computer graphics and animation. Existing methods for human-object interaction often neglect scene context, leading to implausible penetrations, while human-scene interaction approaches struggle to coordinate fine-grained manipulations with long-range navigation. To address these limitations, we propose \textbf{HOSIG}, a novel framework for synthesizing full-body interactions through hierarchical scene perception. Our method decouples the task into three key components: 1) a \textit{scene-aware grasp pose generator} that ensures collision-free whole-body postures with precise hand-object contact by integrating local geometry constraints, 2) a \textit{heuristic navigation algorithm} that autonomously plans obstacle-avoiding paths in complex indoor environments via compressed 2D floor maps and dual-component spatial reasoning, and 3) a \textit{scene-guided motion diffusion model} that generates trajectory-controlled, full-body motions with finger-level accuracy by incorporating spatial anchors and dual-space gradient-based guidance. Extensive experiments on the TRUMANS dataset demonstrate superior performance over state-of-the-art methods. Notably, our framework supports unlimited motion length through autoregressive generation and requires minimal manual intervention. This work bridges the critical gap between scene-aware navigation and dexterous object manipulation, advancing the frontier of embodied interaction synthesis. Project page: \textcolor{magenta}{https://yw0208.github.io/hosig/}.
\end{abstract}

\section{Introduction}

Embodied intelligence represents a pivotal frontier in AI research, aiming to develop agents capable of navigating and manipulating 3D environments. While existing works have extensively explored human-object interaction (HOI)~\cite{taheri2024grip, zhang2024ood, petrov2024tridi, wu2022saga, zheng2023coop, Li2023ObjectMG} and human-scene interaction (HSI)~\cite{karunratanakul2023temos, xiao2023unihsi, zhao2023dimos, zhang2022gamma, wang2024move}, few address the integrated human-object-scene interaction (HOSI) challenge~\cite{wu2024humanlevel, jiang2024trumans, jiang2024autonomous, Lu2024CHOICECH}. As shown in Figure~\ref{fig:teaser}, our goal is to enable characters to move in complex scenes, complete precise object operations, and seamlessly connect them into a long-term motion. The inherent complexity of synthesizing these multimodal interactions presents critical unsolved challenges in spatial reasoning and motion coordination.

While recent advancements in HOI~\cite{yang2024fhoi, song2024hoianimator, peng2023hoidiff, taheri2022goal, ghosh2023imos, Zhang2024ManiDextHM}, they predominantly neglect 3D scene constraints. This oversight leads to implausible human-scene interpenetration, highlighting the necessity for scene-aware reasoning. Besides ignoring the scene in HOI, current HSI approaches exhibit two limitations: 1) Global scene encoding methods~\cite{wang2024move, wang2022humanise, huang2023diffusion} lack granularity for precise motion synthesis, and 2) Local context perception strategies~\cite{cen2024generating, mao2022contact, ghosh2021synthesis} struggle with pathfinding in complex environments, resulting in persistent interpenetration during long-range motion generation. These dual challenges underscore the need for unified scene-object-agent coordination.  

Inspired by the above observations, we propose \textbf{HOSIG}, a hierarchical scene-aware framework that integrates three core components: (1) \textit{Scene-aware grasp pose generator} for precise object manipulation with hand contact in 3D environments, (2) \textit{Collision-aware navigation planner} enabling obstacle-avoiding pathfinding in complex scenes, and (3) \textit{Trajectory-controlled motion synthesizer} generating unrestricted-length whole-body motions through multi-modal condition integration. Our hierarchical architecture operates through three perception levels: \textit{local} (object grasping positions), \textit{global} (scene navigation topology), and \textit{path-aligned} (continuous spatial guidance). Unlike prior works, HOSIG achieves unified coordination of dynamic object manipulation and static scene interaction while maintaining motion coherence through iterative refinement with spatial constraints.  

Our technical implementation features three key innovations. First, the grasp pose generator augments the cVAE framework~\cite{taheri2022goal} with local scene geometry constraints, producing physically-plausible hand-object orientations that prevent scene interpenetration. Second, a novel 2D scene abstraction layer enables efficient navigation through 3D environments via heuristic pathfinding on compressed obstacle-aware maps, dynamically generating obstacle-aware motion graphs. Third, building upon ControlNet's conditioning paradigm~\cite{zhang2023controlnet}, we develop a multi-condition diffusion framework that simultaneously integrates: (a) \textit{spatial anchors} from navigation paths, (b) \textit{fine-grained hand control} through grasp poses, and (c) \textit{path-aligned scene priors} for continuous interaction optimization. This unified architecture achieves finger-level motion precision without auxiliary hand modules, surpassing previous trajectory-control methods through iterative spatial constraint.

In summary, our principal contributions are:
\begin{itemize}
\item[$\bullet$] A scene-geometry constrained grasp generator producing interpenetration-free full-body poses via cVAE augmentation.

\item[$\bullet$] A 2D scene abstraction method with heuristic pathfinding for autonomous obstacle-aware navigation.

\item[$\bullet$] A trajectory-language diffusion model integrating spatial anchors and scene guidance for finger-level motion control.

\item[$\bullet$] An autoregressive HOSI pipeline achieving unlimited-length motion synthesis with full automation.
\end{itemize}

\section{Related Work}

\subsection{Human-Object Interaction Generation}
Research on human-object interaction (HOI) motion generation has progressed through two main paradigms. Early approaches primarily utilized reinforcement learning (RL), with initial works like \cite{peng2019mcp, peng2021amp} achieving basic interactions such as box touching. Subsequent RL methods developed more complex skills including basketball dribbling~\cite{liu2018basketball}, tennis playing~\cite{yuan2023tennis}, multi-object manipulation~\cite{wang2023physhoi}, and box carrying~\cite{hassan2023physicalcharacterscene}. 

The second paradigm employs generative models for direct motion synthesis. GOAL~\cite{taheri2022goal} pioneered this direction using cVAEs for grasping motions. IMoS~\cite{ghosh2023imos} later enabled action-label conditioned generation (e.g., photography, drinking). Recent works focus on language-conditioned generation: OOD-HOI~\cite{zhang2024ood} synthesizes human-object motions separately with contact optimization, while HOI-Diff~\cite{peng2023hoidiff} employs affordance-guided diffusion. Other key innovations include TriDi's unified modeling of bodies/objects/contacts~\cite{petrov2024tridi} and CHOIS's trajectory-constrained synthesis~\cite{li2024chois}. InterDiff~\cite{xu2023interdiff} further enables sequential generation from initial states.

\subsection{Human-Scene Interaction Generation}
Human-scene interaction (HSI) generation focuses on interactions with static environments, distinguished from HOI by the immobility of target objects. We classify scene interactions (e.g., with chairs/sofas) as HSI rather than HOI. Current HSI research bifurcates into two technical directions: (1) scene-aware locomotion and (2) semantic interaction synthesis.

\begin{figure*}[th]
    \centering
    \includegraphics[width=\textwidth]{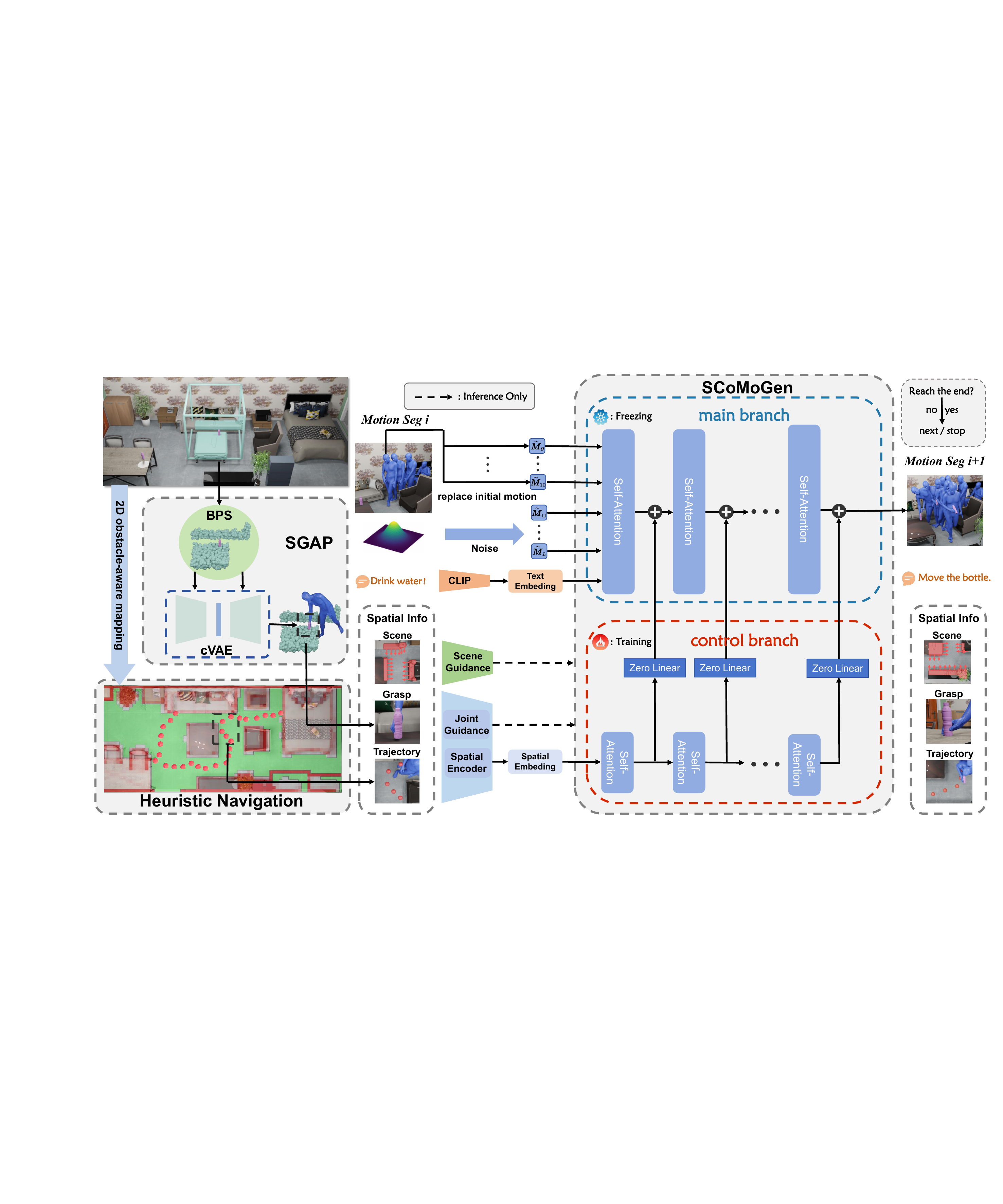}
    \caption{\textbf{Overview of Our Pipeline.} HOSIG can iteratively generate long-term motions based on spatial information, text, and the previous motion clip. There are three parts worth noting in the pipeline: (1) \textbf{SGAP} generates fine grasping postures to ensure the quality of character interaction. (2) \textbf{Heuristic Navigation} generates sparse human root joint trajectories to constrain the subsequent generated motions to be within the traversable area. (3) \textbf{SCoMoGen} uses a dual-branch design to achieve spatial control and adds additional joint \& scene guidance during inference to achieve high-precision control.}
    \label{fig: pipeline}
\end{figure*}

For locomotion generation, existing methods adopt either data-driven~\cite{wang2024pacer+, zhang2022wanderings, rempe2023trace} or algorithmic approaches~\cite{wang2022towards}. While data-driven methods struggle in novel complex environments, our work proposes an algorithmic solution for robust navigation. Semantic interaction synthesis initially produced static poses in scenes~\cite{zhang2020generatingpeopleinscene, li2024genzi, zhao2022coins, xuan2023narrator}, later evolving into language-guided systems~\cite{wang2022humanise, wang2024move, huang2023diffusion} though requiring post-hoc optimization for scene compliance. Recent advances include concurrent navigation-interaction frameworks~\cite{yi2024generatingscenetext} and video-generation-based methods~\cite{li2024zerohsi} with strong generalization but limited contact realism and motion range.

Closest to our approach, \cite{jiang2024trumans, jiang2024autonomous} enable triadic human-object-scene interactions. However, these require frame-level action labels and precise finger positions, whereas our method achieves comparable results using only initial/final object states.

\section{Method}

\subsection{Problem Formulation}

We first formalize the task definition with five essential input parameters: (1) scene $\mathcal{S}$, (2) initial human position $\mathbf{p}_0 \in \mathbb{R}^3$, (3) object start pose $\mathbf{T}_s \in SE(3)$, (4) object target pose $\mathbf{T}_t \in SE(3)$, and (5) object mesh $\mathcal{O}$. Given these parameters, our HOSIG produces synchronized motion trajectories $\mathcal{M} = (\mathcal{M}_h, \mathcal{M}_o)$ containing both human motion $\mathcal{M}_h$ and object motion $\mathcal{M}_o$. A standard interaction sequence typically comprises three phases: initial approach (human navigation to $\mathbf{T}_s$), object manipulation (grasping and transportation from $\mathbf{T}_s$ to $\mathbf{T}_t$), and final placement (precise positioning at $\mathbf{T}_t$). Moreover, our framework supports extension through auxiliary interaction nodes to achieve functions such as sitting on a chair. Please refer to the supplementary materials for more applications.

As shown in Figure~\ref{fig: pipeline}, the HOSIG framework implements its functionality through three interconnected components operating in a collaborative pipeline. The first module, \textbf{S}cene-Aware \textbf{G}r\textbf{a}sp \textbf{P}ose Generation (SGAP), computes feasible full-body grasp poses as anchors for other modules. Subsequently, our novel heuristic navigation algorithm with obstacle avoidance constraints computes collision-free 3D trajectories connecting the anchors. The final component, \textbf{S}cene-Guided \textbf{Co}ntrollable \textbf{Mo}tion \textbf{Gen}eration (SCoMoGen), synthesizes continuous motion sequences along trajectories. The subsequent sections elaborate on each module's technical formulation.

\subsection{Scene-Aware Grasp Pose Generation}
\label{sec:sgap}

As shown in Figure~\ref{fig: pipeline}, the SGAP module is implemented as a conditional variational autoencoder (cVAE) comprising an encoder-decoder architecture. The encoder's input tensor $X \in \mathbb{R}^{d}$ is formed through the concatenation operation:
\begin{equation}
    X=\left[ \varTheta , \beta , \mathbf{V}, \mathbf{D}^{h\rightarrow o}, \hat{\mathbf{h}}, \mathbf{t}^o, \mathbf{B}^o, \mathbf{B}^s \right] 
\end{equation}
where $\varTheta \in \mathbb{R}^{3N_j}$ and $\beta \in \mathbb{R}^{10}$ denote the SMPL-X pose and shape parameters respectively, $\mathbf{V} \in \mathbb{R}^{400\times3}$ represents the sampled body mesh vertices, $\mathbf{D}^{h\rightarrow o} \in \mathbb{R}^{400\times3}$ encodes vertex-wise directional offsets between the human body mesh and the nearest object surface points, $\hat{\mathbf{h}} \in \mathbb{R}^{3}$ specifies the head orientation unit vector, $\mathbf{t}^o \in \mathbb{R}^{3}$ captures the object's translational state, and $\mathbf{B}^o \in \mathbb{R}^{1024}$ denotes the Basis Point Set (BPS) encoding of the object geometry.

\begin{figure}[t]
    \centering
    \includegraphics[width=\linewidth]{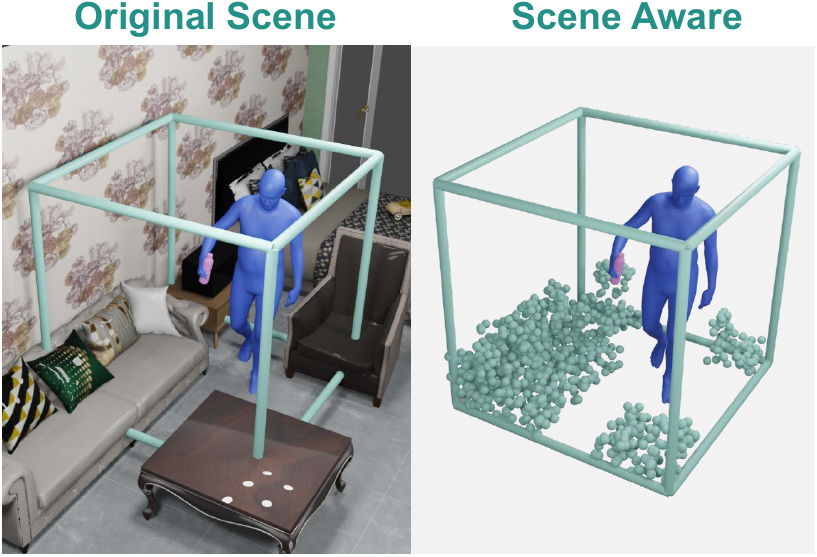}
    \caption{\textbf{Visulization of Scene-Aware in SGAP.} The green box is centered on the purple bottle. In SGAP, only the sparse scene point cloud inside the box is used, as shown in the right figure.}
    \label{fig: scene aware}
\end{figure}

The critical component $\mathbf{B}^s$ constitutes the primary scene perception mechanism in SGAP. As illustrated in Figure~\ref{fig: scene aware}, given a z-up object translation $\mathbf{t}^o = (x, y, z)$, we construct a scene context volume bounded by $[x-0.8, x+0.8] \times [y-0.8, y+0.8] \times [0.2, 1.8]$ meters, forming a $1.6^3 \ m^3$ cubic region. To capture scene geometry, we employ a volumetric sampling strategy that generates a dense point cloud $\mathcal{P}_s$ containing both surface vertices and interior points from the scene mesh. This sampling follows a voxel grid resolution of 8 cm³, ensuring complete spatial coverage. The BPS transformation processes $\mathcal{P}_s$ through basis projections to produce the scene encoding $\mathbf{B}^s \in \mathbb{R}^{1024}$. These scene features condition the pose generation process, enabling synthesis of scene-adapted full-body poses.

To enforce geometric compatibility between human motions and environmental structures, we implement a physics-informed scene distance loss. The scene distance loss $\mathcal{L}_{sd}$ operates on the joint-space representation:
\begin{equation}
    \mathcal{L} _{sd} = -\frac{1}{N}\sum_{j=1}^{22}\sum_{k=1}^{N}  \|\mathbf{J}_j - \mathbf{S}_k\|_2^2
    \label{eq:scene distance}
\end{equation}
where $\mathbf{J}_j \in \mathbb{R}^3$ denotes the $j$-th body joint position from the SMPL-X kinematic tree. $\mathbf{S}_k \in \mathbb{R}^3$ represents the $k$-th point in the localized scene point cloud $\mathcal{P}_s$, as shown in Figure~\ref{fig: scene aware}. Point numbers $N$ is variable, depending on the number of point clouds in the current localized scene. This formulation imposes stronger penalties as joints approach scene surfaces. During training, $\mathcal{L}_{sd}$ backpropagates collision-avoidance constraints through the cVAE's latent space, encouraging the generator to produce poses maintaining clearance from scene geometry while preserving natural motion kinematics. 

During inference, SGAP's output parameters undergo selecting to extract three critical control signals: root joint translation $\mathbf{a}_r \in \mathbb{R}^3$, wrist joint position $\mathbf{a}_w \in \mathbb{R}^3$, and hand joint rotations $\mathbf{a}_h \in \mathbb{R}^{15\times6}$. These elements constitute the spatial anchor tuple $\mathcal{A} = (\mathbf{a}_r, \mathbf{a}_w, \mathbf{a}_h)$, which serves as key constraints for subsequent motion synthesis.

\subsection{Heuristic Navigation on 2D Obstacle-Aware Map}
\label{sec:algorithm}

\begin{figure}[t]
    \centering
    \includegraphics[width=\linewidth]{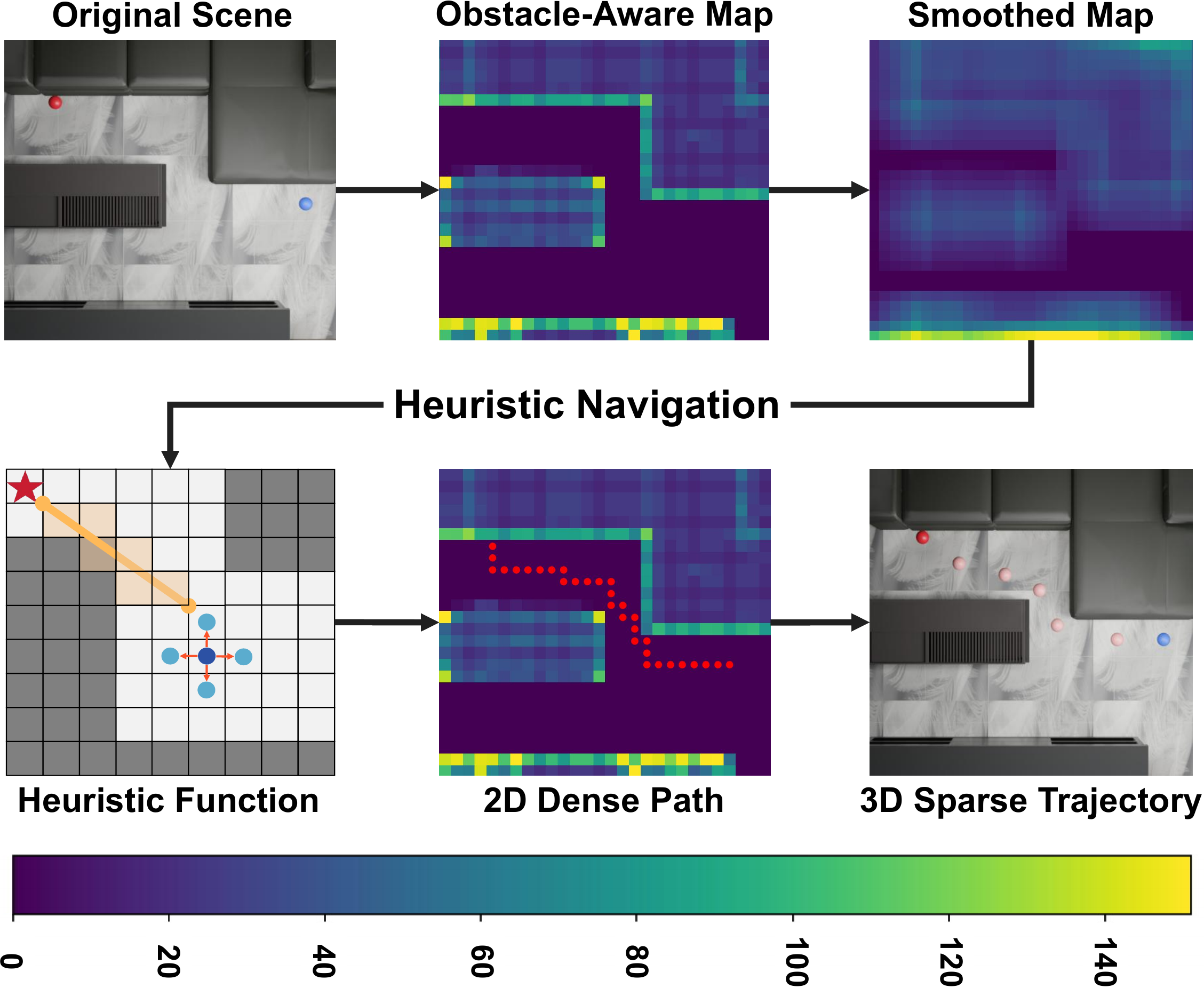}
    \caption{\textbf{Pipeline of Heuristic Navigation.} The blue ball in the original scene is the starting point, and the red ball is the end point. The obstacle-aware map is presented in the form of a heat map, and the values correspond to the axis at the bottom. In \textit{Heuristic Function}, the dark blue dot represents the current node, the light blue dot represents the candidate node, and the red star represents the end point.}
    \label{fig: navigation}
\end{figure}

Our Heuristic Navigation framework constitutes a customized A* variant specifically engineered for 3D point cloud environments, designed as a plug-and-play module compatible with diverse motion generation models. Crucially, the algorithm maintains full replaceability. Any trajectory synthesis method satisfying interface requirements can be integrated into our pipeline. This design intentionally validates our hypothesis: global scene perception substantially outperforms local perception strategies for long-range navigation in complex 3D environments. Here are the steps of our method:

As illustrated in Figure~\ref{fig: navigation}, the pipeline executes four cohesive stages. Initially, the 3D point cloud is partitioned into volumetric blocks and compressed vertically into a 2D heat map, i.e., obstacle-aware map, where each grid cell encodes the cumulative point numbers along the vertical axis. Then, this representation undergoes convolution smoothing to deliberately blur boundaries between obstacles and walkable regions, mitigating the algorithm's tendency to generate trajectories adhering close to obstacle surfaces. Subsequently, a dense 2D path is computed via our implementation, employing a dual-component heuristic function. As shown in the lower left corner of the Figure~\ref{fig: navigation}, one term evaluates Euclidean distance to the goal, like the orange line. While the other term utilizes the Bresenham line algorithm to compute cumulative traversal costs by summing values along the direct path between candidate points and the destination, like the light orange area. Finally, the resulting 2D dense path is adaptively downsampled, added height, and transformed back into the native 3D coordinate system to yield a 3D sparse trajectory. This trajectory will be added to the $\mathbf{a}_r$ of the spatial anchor tuple $\mathcal{A}$ as a root joint control signal to serve the subsequent controllable motion generation.

Due to space limitations, if you are curious about the details of this algorithm, please refer to the detailed description in the supplementary materials.

\subsection{Scene-Guided Controllable Motion Generation}
\label{sec: sgcmg}

To enable the generation of high-fidelity interactions, two key challenges must be addressed. First, effective control over human body movement along the specified trajectory is required, with particular emphasis on enabling precise hand movements for object grasping. Second, in narrow and complex scenes, relying solely on the trajectory is insufficient to guarantee scene avoidance. Thus, a mechanism for deeply integrating scene information into the motion generation process is necessary. To address these two challenges, we respectively propose two corresponding approaches.

\textbf{Controllable Generation} A ControlNet-like architecture is employed, where spatial control signals serve as inputs to the control branch for generating motions aligned with specified spatial trajectories, as illustrated in Figure \ref{fig: pipeline}. Specifically, the main branch of SCoMoGen takes the motion noises $\left\{ \tilde{M}_t \right\} _{t=0}^{L}$ and the text embedding encoded by CLIP as inputs, and outputs next motion segmentation. Among them, for the coherence of motions, $\tilde{M}_0$ to $\tilde{M}_{10}$ are the last 11 frames of the previous motion segment. The control branch of SCoMoGen takes control signals as inputs. The control signals are converted from spatial anchor tuples $\mathcal{A}$, where the uncontrolled parts are set to 0. Key design elements consist of initializing the control branch with pretrained weights from the main branch and implementing zero-initialized connection layers. These design enable the incorporation of additional control capabilities while preserving the original model's generative performance. Notably, for the first time, our proposed SCoMoGen achieves full-body control encompassing hand control. This advancement stems from optimized hand pose representation: wrist and 15 finger joints are parameterized using 6D global rotations within the world coordinate system, contrasting with SMPL-X's rotations relative to parent joints. Deviations in trunk joint rotations result in cumulative error propagation through kinematic chains in conventional methods. The proposed representation determines hand poses through 16 global joint rotations combined with wrist positioning, effectively eliminating error accumulation pathways.

\begin{figure}[th]
    \centering
    \includegraphics[width=\linewidth]{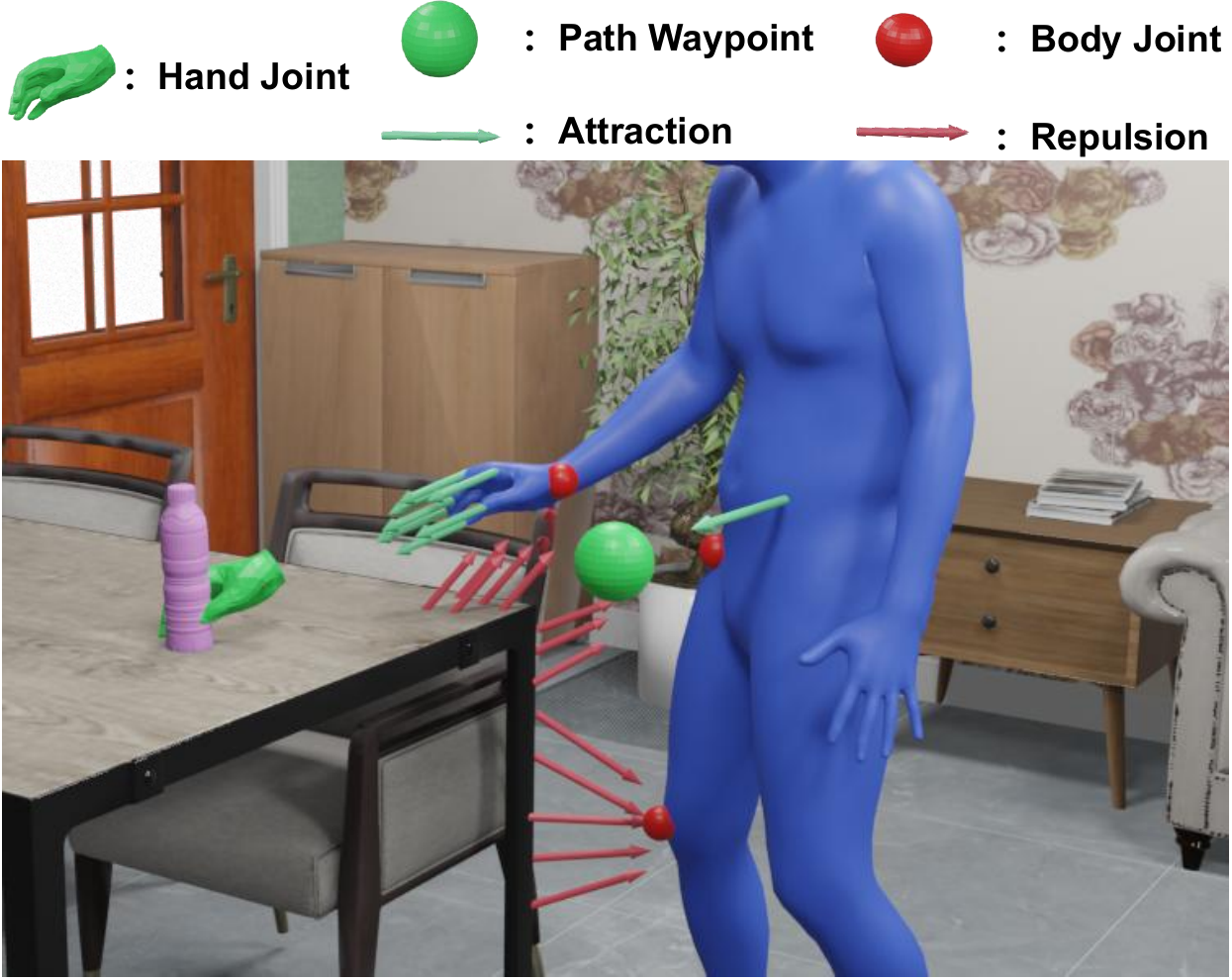}
    \caption{\textbf{Visulization of Guidance in SCoMoGen.} Precise control of motions is achieved through gradient-based guidance. Green arrows represent joints being attracted by green anchors (hand joints, path waypoints). Red arrows represent the repulsive force on the joint (red body joint) close to the scene.}
    \label{fig: scene guidance}
\end{figure}

\textbf{Joint \& Scene Guidance} Despite the implemented spatial control mechanisms, generating motions that satisfy precision requirements remains challenging, since minor deviations in interactive tasks can produce noticeable motion artifacts. To address this limitation, gradient-based guidance~\cite{guo2024gradient} is incorporated to facilitate human-scene interaction generation without requiring additional motion post-processing. As depicted in Figure~\ref{fig: scene guidance}, two complementary constraints joint-level and scene-level guidance are implemented within the framework.

The joint constraint $\mathcal{L}_j$ establishes attraction forces between body joints $\mathcal{J}$ and target anchors $\mathcal{A} = (\mathbf{a}_r, \mathbf{a}_w, \mathbf{a}_h)$ in 3D space. For root joint $j_{r}$ and root anchors $\mathbf{a}_r$, the loss is formulated as:
\begin{equation}
\mathcal{L}_j^{r} = \sum_{i=1}^{L_{r}} \| j_{r}^{(i)} - \mathbf{a}_{r}^{(i)} \|_2^2
\end{equation}
where $L_r$ is the number of path waypoints.
For hand-object interactions, the constraint is implemented using:

\begin{equation}
\mathcal{L} _{j}^{hand}=\sum_{i=1}^{L_{h}}{\parallel j_{w}^{(i)}}-\mathbf{a}_{w}^{\left( i \right)}\parallel _{2}^{2}+\sum_{i=1}^{L_{h}}{\parallel r_{h}^{(i)}}-\mathbf{a}_{h}^{\left( i \right)}\parallel _{2}^{2}
\end{equation}
where $j_{w}^{(i)}$ denotes wrist positions and $r_{h}^{(i)}$ represents 16 hand rotations. The parameter $L_h$ typically equals 2, corresponding to two key poses generated by SGAP for object pickup and placement. This parameter can be extended depending on application requirements.

The scene constraint $\mathcal{L}_s$ generates adaptive repulsion forces through path-aligned point cloud. For each joint $j$, the repulsion loss is defined as:

\begin{equation}
\mathcal{L}_s = -\sum_{j \in \mathcal{J}} \sum_{p \in \mathcal{N}(j,\ 0.3\text{m})} \|j - p\|_2^2
\label{eq: scene guidance}
\end{equation}
where $\mathcal{N}(j, r)$  represents scene points located within a radius $r$ of joint $j$. By restricting attention to the local $\mathcal{N}(j, r)$, irrelevant scene elements are excluded, sharply reducing computational overhead and steering optimization toward a more precise direction.

\begin{table*}[t]
\small
\setlength{\tabcolsep}{5.4pt}
\begin{tabular}{cccccccccc}
\hline
\multirow{2}{*}{Methods} & \multicolumn{3}{c}{Object Locomotion}      & \multicolumn{3}{c}{Scene Interaction}    & \multicolumn{3}{c}{Object Interaction}   \\ \cline{2-4} \cline{5-7} \cline{8-10}
                         & Dist. $\downarrow$          & Time $\downarrow$    & Rate $\uparrow$           & Pene. Rate $\downarrow$     & Pene. Mean $\downarrow$ & Pene. Max $\downarrow$ & Contact Rate $\uparrow$  & Pene. Mean $\downarrow$ & Pene. Max $\downarrow$ \\ \hline
LINGO                    & 0.4169         & 16.6100   & 0.2333          & 0.3175          & 0.3745     & 89.5247   & 0.2418          & 0.0001     & 0.0070    \\
CHOIS                    & 0.3602         & 11.5033 & 0.0333          & 0.4134          & 26.3719    & 4271.7520 & 0.2809          & 0.0001     & 0.0056    \\ \hline
Ours                     & \textbf{0.0270} & 13.0367 & \textbf{0.9333} & \textbf{0.1851} & 0.6562     & 201.8264  & \textbf{0.9800} & 0.0007     & 0.0113    \\ \hline
\end{tabular}
\caption{\textbf{Quantitative results of human-object-scene interaction generation.} This mainly involves the interaction between human and scenes, human and objects when characters operate objects in the scene. At the same time, this also evaluates the efficiency and accuracy of the character in carrying objects.}
\label{tab:main}
\end{table*}

The described losses are not directly applied for result optimization, but rather integrated with the diffusion framework to refine the predicted mean $\mu _t$ at designated timesteps $t$ through: 
\begin{equation}
\mu _t=\mu _t-\tau \nabla _{\mu _t}\left( \lambda _1\mathcal{L} _{j}^{root}+\lambda _2\mathcal{L} _{j}^{hand}+\lambda _3\mathcal{L} _{s}^{} \right)
\end{equation}
where $\tau$ controls the optimization magnitude and $\lambda_{1,2,3}$ balance constraint contributions. This approach demonstrates superior motion naturalness compared to direct iterative result optimization by preventing artifact generation from over-constrained objectives.

\section{Experiments}

\subsection{Implementation Details}
Our evaluation protocol comprehensively assesses both absolute performance and component effectiveness. All experiments are conducted on the TRUMANS dataset~\cite{jiang2024trumans}, containing 100 indoor scenes with annotated human-object interactions. We establish two evaluation axes: 1) comparison against SOTA methods in HOI and HSI generation, and 2) ablation studies isolating our core technical innovations. 

\textbf{Metrics} 
We employ two complementary metric categories: Object Locomotion Assessment evaluates spatial-temporal performance through: (1) terminal positioning accuracy (Dist: Euclidean distance to target), (2) temporal efficiency (Time: task duration), and (3) reliability (Rate: success proportion within 0.05m threshold). Human Interaction Analysis quantifies physical plausibility via SDF-based penetration metrics: collision frequency (Penetration Rate), hand-object contact quality (Contact Rate), average severity (Mean Penetration Volume), and worst-case failures (Max Penetration Depth).

\textbf{SOTA Comparisons}
While no existing method achieves identical functionality, we select two representative baselines LINGO~\cite{jiang2024autonomous} and CHOIS~\cite{shi2023controllable} for fair comparison. 
Quantitative evaluations cover three axes: object locomotion, human-scene interaction, hand-object interaction. The user study employs 30 sets of comparison videos assessing generation performance. Detailed metric explanation, visual comparisons, and more implementation details like baseline reproduction and training details, are provided in the supplementary material.

\textbf{Ablation Study}
We validate our three key designs through controlled experiments:
(1) \textbf{Scene-Aware}: Removes local scene information $B^s$ and relative scene distance loss function $\mathcal{L}_{sd}$.
(2) \textbf{Heuristic Navigation}: No additional experiments were set up, and its efficiency can be reflected in Table~\ref{tab:main}.
(3) \textbf{Scene-Guided}: Disables gradient-based guidance (Eq.~\ref{eq: scene guidance}), relying solely on joint-based optimization.
\subsection{Results}

\subsubsection{Comparison}

\textbf{Quantitative Experiments} Our method demonstrates \textbf{three key advantages} through comprehensive benchmarking. As shown in Table~\ref{tab:main}, HOSIG achieves a 93.3\% success rate in object locomotion tasks, outperforming LINGO (23.3\%) and CHOIS (3.3\%) by factors of 4.0$\times$ and 28.3$\times$ respectively. This quantifies our method's enhanced controllability in full-body motion generation, particularly in complex navigation-objective coordination scenarios. The hierarchical scene perception mechanism yields a penetration rate of 42\% - 55\% lower than baselines. While LINGO shows lower mean penetration volume (0.37 vs 0.66), our method prevents catastrophic failures as evidenced by maximum penetration values: 201.8 vs CHOIS' 4,271.8. This confirms our layered constraint system effectively balances micro/macro scene interactions. As comparable contact rates (98\% vs 24-28\%), quantitative analysis reveals our full-body generation enables \textit{functional} hand-object interactions, not just torching as baselines. Although baselines have low metrics on the penetration volume, this is mainly due to their extremely low hand-object contact rate. Overall, ours achieves better full-body HOSI generation.


\begin{figure}[h]
    \centering
    \includegraphics[width=\linewidth]{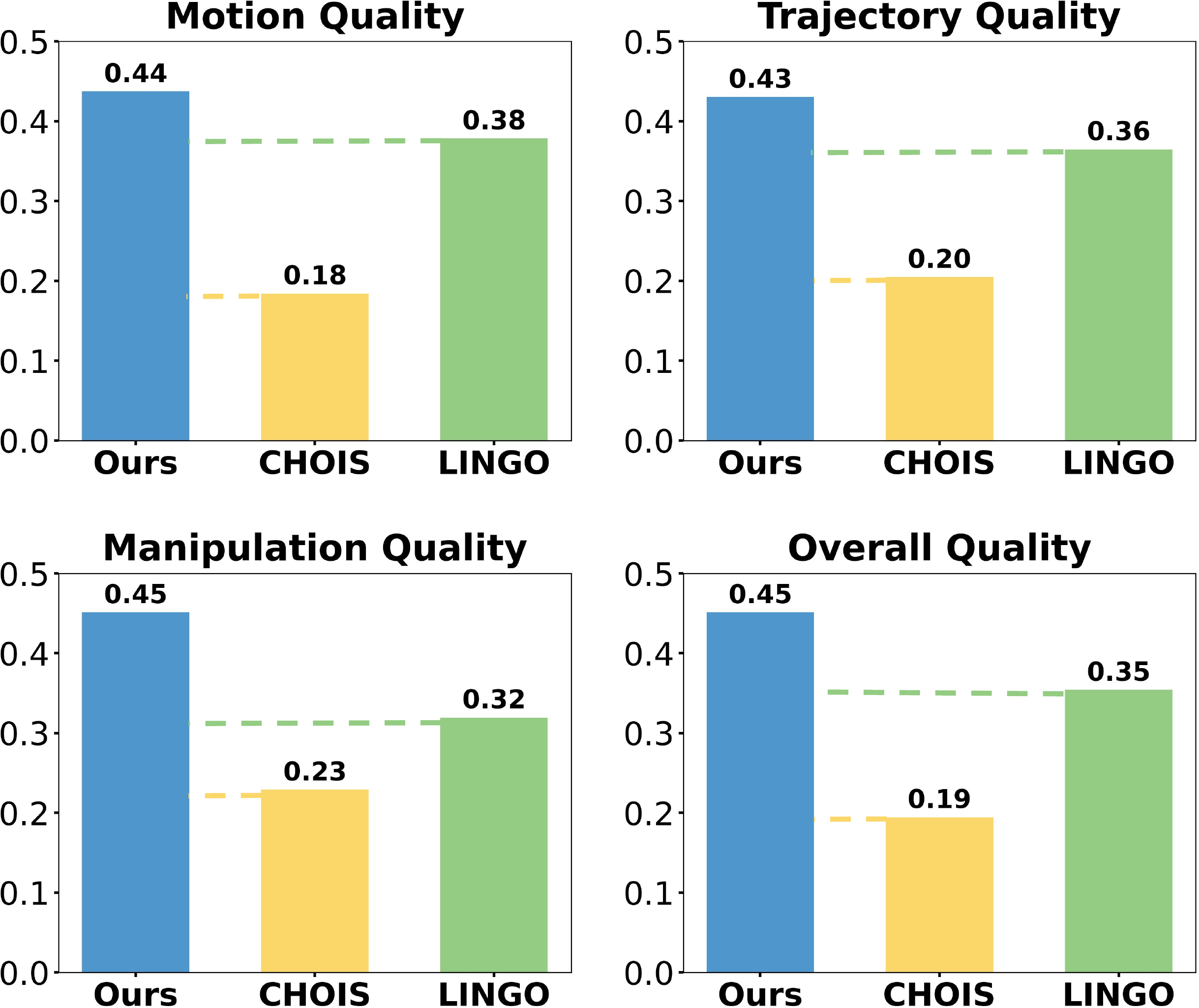}
    \caption{\textbf{User Study.} Users are asked to rate generated motions based on four different indicators, with the best being 3 points, the middle being 2 points, and the worst being 1 point. The scores are then tallied and the proportion of each method is calculated. The best result that could be obtained is 0.50. The closer it is to 0.50, the better the performance.}
    \label{fig: user study}
\end{figure}

\textbf{User Study} The user study involved participants evaluating motion generation outcomes through four quality metrics: Motion Quality, Trajectory Quality, Manipulation Quality, and Overall Quality. Performance assessment is conducted by measuring the relative contribution of each method to the aggregated score, with metric definitions provided in supplementary materials. As depicted in Figure~\ref{fig: user study}, our HOSIG method outperforms both baselines across all metrics, achieving values approaching 50\% – the theoretical maximum for each indicator. This demonstrates that our method is consistently preferred over alternatives in pairwise comparisons. The subjective evaluations align with quantitative experimental results, closely matching the observed performance hierarchy HOSIG $>$ LINGO $>$ CHOIS. These findings collectively demonstrate that hierarchical scene perception enables HOSIG to achieve robust human-object-scene interaction synthesis.

\subsubsection{Ablation Study}

\

\textbf{Scene-Aware} The Scene-Aware variant (removing local point cloud constraints) reveals critical insights into our spatial reasoning mechanism. As shown in Table~\ref{tab: able sa}, the full SGAP model improves scene interaction metrics by 16.3\% (penetration rate), 69.9\% (mean penetration volume), and 59.4\% (max penetration volume) compared to the ablated version. Crucially, object interaction precision remained stable with $<2\%$ variation in grasp success rate, confirming that scene awareness enhances environment adaptation without compromising manipulation capabilities. Further analysis demonstrated that models trained without the scene distance loss (Eq.~\ref{eq:scene distance}) suffer from higher penetration in scene avoidance, as unstructured scene features introduce noise in the latent space. In conclusion, SGAP not only needs additional scene information $B^s$ to perceive the scene, but also needs corresponding loss $\mathcal{L}_{sd}$ to learn how to use the scene information. Both scene information and scene distance loss are indispensable.

\textbf{Heuristic Navigation} The efficacy of our Heuristic Navigation framework manifests most prominently in two critical metrics: temporal efficiency (\textbf{Time} in Object Locomotion) and collision integrity (\textbf{Penetration Rate} in Scene Interaction). As quantified in Table~\ref{tab:main}, our approach achieves a $3.6$ second average improvement in task completion time over the LINGO baseline. This significant acceleration stems directly from global scene comprehension, which enables anticipatory obstacle avoidance and optimal route planning. Conversely, local perception methods frequently encounter pathological scenarios, particularly in geometrically complex environments. Myopic decision making leads to navigation dead-ends and recovery behaviors, as qualitatively demonstrated in our supplementary visualizations. Furthermore, our method's superior \textit{Penetration Rate} ($\downarrow 23\%$ versus LINGO) also confirms enhanced path precision.

\begin{table}[th]
\small
\begin{tabular}{cccc}
\hline
\multirow{2}{*}{Methods} & \multicolumn{3}{c}{Scene Interaction}  \\ \cline{2-4} 
                         & Pene. Rate  $\downarrow$   & Pene. Mean  $\downarrow$ & Pene. Max  $\downarrow$ \\ \hline
SGAP                     & \textbf{0.5533}        & \textbf{0.4484}     & \textbf{12.0956}   \\
w/o sd                   & 0.7037        & 3.4615     & 134.6292  \\
w/o sa             & 0.6611        & 1.4918     & 29.8254   \\ \hline
\multirow{2}{*}{Methods} & \multicolumn{3}{c}{Object Interaction} \\ \cline{2-4} 
                         & Contact. Rate  $\uparrow$ & Pene. Mean  $\downarrow$ & Pene. Max  $\downarrow$ \\ \hline
SGAP                     & 0.9833        & 0.0011     & 0.0071    \\
w/o sd                   & 0.9815        & \textbf{0.0007}     & \textbf{0.0053}    \\
w/o sa             & \textbf{1.0000}        & \textbf{0.0007}     & 0.0061    \\ \hline
\end{tabular}
\caption{\textbf{Ablation studies of SGAP.} The w/o sd means training SGAP without scene distance loss. The w/o sa means completely removing scene perception, including scene distance loss constraints and scene information $B^s$ input to the model.}
\label{tab: able sa}
\end{table}

\begin{table}[h]
\small
\begin{tabular}{cccc}
\hline
\multirow{2}{*}{Methods} & \multicolumn{3}{c}{Object Locomotion}   \\ \cline{2-4} 
                         & Dist $\downarrow$         & Time $\downarrow$      & Rate $\uparrow$      \\ \hline
SCoMoGen                 & \textbf{0.0270}        & 13.0367    & \textbf{0.9333}     \\
w/o sg                   & 0.5540        & 12.6511    & 0.6667     \\ \hline
\multirow{2}{*}{Methods} & \multicolumn{3}{c}{Scene Interaction}   \\ \cline{2-4} 
                         & Pene. Rate $\downarrow$   & Pene. Mean $\downarrow$ & Pene. Max $\downarrow$  \\ \hline
SCoMoGen                 & \textbf{0.1851}        & \textbf{0.6562}     & \textbf{201.8264}   \\
w/o sg                   & 0.3043        & 336.6398   & 27326.6445 \\ \hline
\multirow{2}{*}{Methods} & \multicolumn{3}{c}{Object Interaction}  \\ \cline{2-4} 
                         & Contact. Rate $\uparrow$ & Pene. Mean $\downarrow$ & Pene. Max $\downarrow$ \\ \hline
SCoMoGen                 & \textbf{0.9800}        & \textbf{0.0007}     & 0.0113     \\
w/o sg                   & 0.9433        & 0.0008     & 0.0083     \\ \hline
\end{tabular}
\caption{\textbf{Ablation studies of SCoMoGen.} The w/o sg means inferencing without scene guidance $\mathcal{L_s}$.}
\label{tab: able sg}
\end{table}

\textbf{Scene-Guided} The Scene-Guided configuration (disabling Eq.~\ref{eq: scene guidance} in gradient-based guidance) exhibits significant performance degradation across all scene interaction metrics. As shown in Table~\ref{tab: able sg}, quantitative results show higher penetration volumes and lower object locomotion stability compared to our full model. Notably, the guidance mechanism improves hand-object alignment precision. Experiments show that scenes are worth using as an additional modality to help generate motions. This is not only to avoid the penetration of human and scenes, but also to improve the quality of humans' activities in the scene, such as manipulating objects.



\section{Conclusions}
We present \textbf{HOSIG}, a hierarchical framework for synthesizing high-fidelity full-body human-object-scene interactions in complex 3D environments. By decoupling the task into scene-aware grasp pose generation, heuristic navigation planning, and scene-guided controllable motion synthesis, our method addresses critical limitations in existing HOI and HSI approaches. The proposed hierarchical scene perception mechanism combines local geometry constraints, compressed 2D obstacle-aware maps, and dual-space diffusion guidance. It ensures collision-free interactions while maintaining precise hand-object contact and natural locomotion. Notably, our framework achieves unlimited motion length through autoregressive generation and requires minimal manual intervention, making it practical for applications in VR, robotics, and animation.

\bibliography{AnonymousSubmission/LaTeX/aaai2026}

\clearpage

\section{Evaluation Metrics}
\label{sec:metrics}

\subsection{Object Locomotion Assessment}
The spatial-temporal performance of object manipulation is quantified through three core metrics. 

\textbf{Dist.} The \textit{Distance} metric measures the Euclidean distance between the object's final position $p_{\text{final}}$ and the target location $p_{\text{target}}$, expressed as $\|p_{\text{final}} - p_{\text{target}}\|_2$ in meters. This metric directly reflects the system's terminal positioning accuracy and spatial control capability, where smaller values indicate better alignment with the target configuration.

\textbf{Time} The \textit{Time} metric captures the temporal efficiency of task execution, calculated as $t_{\text{end}} - t_{\text{start}}$ in seconds. It evaluates the overall planning optimality and motion coordination effectiveness, with lower values representing faster task completion without unnecessary detours or oscillations.

\textbf{Rate} The \textit{Success Rate} provides a threshold-based binary evaluation, defined as the proportion of trials where the final distance falls within 0.05 meters: $\frac{1}{T}\sum_{t=1}^T \mathbb{I}(\|p_{\text{final}}^{(t)} - p_{\text{target}}\|_2 \leq 0.05)$. This metric serves as a comprehensive indicator of system reliability, combining spatial precision and temporal consistency requirements.

\subsection{Human Interaction Analysis} 
To assess the physical plausibility of human-scene and human-object interactions, we introduce penetration metrics based on signed distance fields (SDF). 

\textbf{Pene. / Contact Rate} The \textit{Penetration Rate} quantifies the frequency of invalid collisions, computed as the temporal average of penetration occurrence: 
\begin{equation}
    \mathcal{R}=\frac{1}{T}\sum_{t=1}^T \mathbb{I}(\exists \mathbf{v} \in \mathcal{V}_h^t: \phi(\mathbf{v}, M) < 0)
\end{equation}
where $\mathcal{V}_h$ is the human vertices, $M$ is the scene/object mesh and $\phi(\cdot)$ is SDF function, $T$ is the total number of human poses. The Penetration Rate measure reveals the system's ability to maintain basic collision avoidance constraints during dynamic interactions. As for \textit{Contact Rate}, it evaluates the quality of hand-object interaction. With a higher contact rate, the object is less likely to be suspended in the air, which means that the motion of manipulating the object is more physically possible.

\textbf{Pene. Mean} The \textit{Mean Penetration Volume} evaluates the spatial extent of penetration artifacts through depth-weighted integration: 
\begin{equation}
    \mathcal{V}_{mean}=\frac{1}{T}\sum_{t=1}^T \sum_{\mathbf{v} \in \mathcal{V}_h^t} |\phi(\mathbf{v}, M)| \cdot \mathbb{I}(\phi(\mathbf{v}, O) < 0)
\end{equation}
This metric characterizes the average penetration severity, with lower values indicating better contact modeling fidelity. 

\textbf{Pene. Max} The \textit{Maximum Penetration Depth} identifies worst-case penetration scenarios:
\begin{equation}
    \mathcal{V}_{max}=\max_{t \in [1, T] } \left[\sum_{\mathbf{v} \in \mathcal{V}_h^t} |\phi(\mathbf{v}, M)| \cdot \mathbb{I}(\phi(\mathbf{v}, O) < 0) \right]
\end{equation}
This conservative measure highlights local failures in collision handling that may lead to visually implausible interactions, even when average penetration volumes appear acceptable.

\subsection{User Study}

Four dimensions are established for perceptual assessment:

\textbf{Motion Quality} evaluates the naturalness of human body movements in isolation. This metric focuses on biomechanical plausibility of limb motions and posture transitions during object manipulation, specifically assessing: (1) natural walking gaits, (2) physically valid grasping/releasing gestures, and (3) absence of unnatural body contortions. Scene context and object dynamics are visually masked during evaluation.

\textbf{Trajectory Quality} measures the spatial rationality of human navigation paths. Evaluators analyze 2D trajectory plots to assess: (1) path optimality (minimal detours), (2) motion smoothness (absence of abrupt turns), and (3) obstacle avoidance capability. Temporal aspects are excluded by using static trajectory visualizations.

\textbf{Manipulation Quality} focuses on human-object physical interaction validity. This criterion examines: (1) hand-object contact consistency during grasping, (2) object stability during transportation (no penetration or unrealistic oscillations), and (3) smooth object motion transitions. Finger-level details are ignored to ensure fair comparison, since two baseline methods can not generate hand poses.

\textbf{Overall Quality} provides integrated assessment of the complete motion synthesis system. Evaluators consider the synergistic effects of: (1) coordination between body motion and navigation trajectory, (2) physical plausibility of combined human-scene-object interactions, and (3) holistic visual realism of the generated motions.

\section{Implementation Details}

This section delineates the critical implementation specifics that bridge our methodological framework with practical execution. The architecture exposition progresses through three coherent dimensions: (1) Systemic orchestration of the core computational pipeline, highlighting the synergistic interaction between principal components and stability-enhancing design choices; (2) Detailed exposition of training protocols encompassing parameter initialization strategies, optimization dynamics, and convergence monitoring mechanisms; (3) Comprehensive description of experimental configurations including dataset curation procedures, and benchmarking methodologies. Collectively, these implementation facets establish the technical foundation for reproducible empirical validation of our proposed framework.

\subsection{Heuristic Navigation Algorithm Particulars}
Unlike learning-based methods that require extensive training data~\cite{yi2024generatingscenetext}, Algorithm~\ref{alg: astar_path} operates on 3D scene pointcloud $\mathcal{S}_p$ while maintaining computational efficiency. Next, we will introduce several key points in the algorithm.

The 3D-to-2D projection converts a whole scene point cloud $S_p$ into a 2D navigable obstacle-aware map $\mathcal{S}_{map}=\{v_{ij}\}$ via vertical voxel quantization. Let $G = \{g_{ij}\}$ denote the 2D grid where each cell $g_{ij}$ corresponds to a $0.1\text{m} \times 0.1\text{m}$ planar region. The cumulative point number $v_{ij}$ of $g_{ij}$ is computed as:
\begin{equation}
v_{ij} = \sum_{(x,y,z) \in S_p} \mathbb{I}((x,y) \in g_{ij}\ \land \ 0.2<z<2)
\label{eq: 2d obstacle-aware map}
\end{equation}
where $\mathbb{I}$ is the indicator function. In short, Equation \ref{eq: 2d obstacle-aware map} is to calculate the number of scene mesh vertices within the range of $g_{ij}$. As for $0.2<z<2$, this is to avoid counting useless information such as the floor and ceiling. This can effectively prevent the character from walking against obstacles. This vertical point density encoding effectively captures obstacles while preserving walkable surfaces, $v_{ij}>0$ means obstacles and $v_{ij}=0$ means walkable. As for trajectory keypoints $\{P_i\}_{i=0}^{n}$,  they usually include preset starting points, ending points, and the root joint $\mathbf{a}_r$ generated by SGAP. During the initialization phase, trajectory keypoints are sequentially paired to form multiple node pairs. Meanwhile, normalization is performed to align the coordinates with  $\mathcal{S}_{map}$.

Our dual-component heuristic $\mathcal{F}$ combines geometric and density awareness. For any two nodes $a=(x_a,y_a)$ and $b=(x_b,y_b)$, we define:

\begin{equation}
\mathcal{F}(a,b) = \underbrace{\|a - b\|_2}_{\text{Euclidean term}} + \lambda \sum_{g \in \mathcal{L}(a,b)} v_g 
\label{eq:heuristic}
\end{equation}
where $\mathcal{L}(a,b)$ denotes the Bresenham line algorithm's output between $a$ and $b$, and $\lambda$ controls obstacle avoidance strength. This formulation guides the search away from obstacle areas while maintaining acceptable arrival efficiency.

As shown in Algorithm~\ref{alg: astar_path}, the ``new path is better'' condition evaluates both accumulated cost and future estimates. The $node$ is a 2D coordinate $(i, j)$. Let $\mathcal{G}(node^{q})$ denote the current path cost from $node^s$ to node $node^q$. A candidate neighbor $node^{q'}$ replaces existing entries in $\mathcal{G}$ if:

\begin{equation}
\mathcal{G}(q) + v_{node^{q'}} < \mathcal{G}(node^{q'}) 
\label{eq:update}
\end{equation}
which means a better path (from $node^s$ to $node^q$ to $node^{q'}$) is found. Meanwhile, make $\mathcal{G}(node^{q'})=\mathcal{G}(q) + v_{node^{q'}}$, $\mathcal{F}$ records the heuristic value from $node^{q'}$ to the end node $node^e$, $\mathcal{C}$ updates $\mathcal{C}(node^{q'})=node^q$ and $\mathcal{Q}$ updates $\{node^{q}:\mathcal{G}(node^{q})+\mathcal{F}(node^{q})\}$. In the next iteration, current node $node^q$ is the node with the minimal value, because $\mathcal{Q}$ is a \textbf{priority queue}. Just keep going forward until reaching the end point $node^e$. When the $node^q$ is the end point, trace back the path from the came-from table $\mathcal{C}$.

In actual use, we can predefine or generate (i.e., $\mathbf{a}_r$ generated by SGAP) any number of ordered nodes, and our algorithm can obtain a feasible path to travel through the scene. For convenience, the path is written as $\mathbf{a}_r$ just like $\mathbf{a}_r$ generated by SGAP.

\begin{algorithm}[t]
\SetAlgoNoLine
\KwIn{3D scene point cloud $\mathcal{S}_{p}$, trajectory keypoints $\{P_i\}_{i=0}^{n}$.}
\KwOut{Optimized sparse path $\{path_i\}_{i=0}^{n-1}$.}

\textbf{Initialize}: Smooth 2D obstacle-aware map $\mathcal{S}_{map} \leftarrow \mathcal{S}_{p}$,  normalized start/goal node pairs $\{node^{s}_{i},node^{e}_{i}\}_{i=0}^{n-1} \leftarrow \{P_i\}_{i=0}^{n}$. \

\For{each $(node^{s}_{i},node^{e}_{i})$ in $\{node^{s}_{i},node^{e}_{i}\}_{i=0}^{n-1}$}{

Initialize priority queue $\mathcal{Q}$, g-score table $\mathcal{G}$, came-from table $\mathcal{C}$; \

Compute f-score $\mathcal{F}(node^{s}_{i}) \leftarrow heuristic(node^{s}_{i},\ node^{e}_{i})$; \

Add $\{node^{s}_{i}:\mathcal{G}(node^{s}_{i})+\mathcal{F}(node^{s}_{i})\}$ to $\mathcal{Q}$; \

\While{$\mathcal{Q}$ is not empty}{
Extract current node $node^q$ with minimal value from $\mathcal{Q}$; \

\If{$current=goal$}{
Backtrack and record dense path $path$; \

break;
}

\For{4-directional $neighbors$ of $node^q$}
{
\If{new path is better}{
Update $\mathcal{G}$, $\mathcal{F}$, $\mathcal{C}$, and push to $\mathcal{Q}$;
}}
}

Adjust path height and coordinates to original space; \

Save $path_i$;}
\caption{Heuristic Navigation Algorithm}
\label{alg: astar_path}
\end{algorithm}

\subsection{Pipeline}
\subsubsection{Scene-Aware Grasp Pose Generation}
The grasp pose generation module produces 10 candidate poses per object-scene configuration. To address implausible grasps caused by interpenetration between the character and the scene, we implement a collision-aware filtering mechanism. Each candidate pose undergoes collision detection against all scene elements. The pose with minimal aggregate collision volume is selected for subsequent processing. This mechanism effectively eliminates physically inconsistent grasps while maintaining sufficient pose diversity.

\subsubsection{Heuristic Navigation Adjustment}
For 2D obstacle-aware navigation, we enforce strict node validity checks during path planning. Any target node falling inside obstacle regions or causing pathfinding failures triggers an interactive adjustment protocol. The system visually overlays obstacle heatmaps on the navigation interface and allows human operators to reposition invalid nodes. This human-in-the-loop correction ensures all navigation targets reside in collision-free regions before executing path planning algorithms. 

Additionally, when utilizing the obtained root joint control signal $\mathbf{a}_r$, we typically apply a control signal every 38 frames. However, this interval is adjustable. To increase the walking speed, the frame interval can be decreased. Conversely, to slow down the walking speed, the frame interval can be increased. In particular, when it comes to performing specific skills such as sitting down and standing up, we only need to additionally apply some root joint control signals with adjusted heights to achieve this.

\subsubsection{Motion Generation Constraints}
The motion synthesis module employs three key constraints for temporal coherence:

1) \textbf{Finger articulation constraints}: Maintain fixed finger joint angles within 10 frames (5 frames before/after grasp events) to ensure stable object manipulation.

2) \textbf{Motion continuity}: Enforce 10-frame overlap between consecutive motion segments, using the final pose of the preceding motion as the initial condition for subsequent motion generation.

3) \textbf{Object trajectory synchronization}: Gradually transition object poses relative to the character hand during transportation phases. This smooth interpolation bridges the discrete “pick up” and “put down” configurations through spherical linear interpolation, preventing abrupt object position jumps.

\subsection{Training Details}

Our three-stage training pipeline utilizes the Trumans~\cite{jiang2024trumans} dataset on a single NVIDIA RTX 3090/4090 GPU. Phase 1 trains the SGAP for 8 hours using the standard $\mathcal{L}_{VAE}$ loss and collision penalty terms. Following MDM, as a diffusion model, phase 2 trains the SCoMoGen's main branch over 48 hours. Note SCoMoGen is trained on TRUMANS totally, and does not use any base models or other datasets.

Phase 3 strictly follows ControlNet's paradigm: 1) Duplicate main branch weights to initialize the control branch. 2) Zero-initialize trainable control layers. 3) Freeze main network while training only control-specific parameters through 24-hour fine-tuning. This strategy injects spatial control signals via residual connections, optimized with $\mathcal{L}_{diff}$ plus $\mathcal{L}_{ctrl} = \|c_{target} - c_{pred}\|_2$ where $c$ denotes control embeddings. Our implementation builds on these open-source works~\cite{taheri2022goal, li2024chois, xie2023omnicontrol}. And we will also make our code public as soon as possible after publication.

\subsection{Experiment Setup}
\textbf{Evaluation of HOSI Generation} The evaluation protocol assesses two critical capabilities: object-agnostic grasping proficiency and scene generalization. We select three daily objects (bottle, cup, mouse) and 10 scenes excluded from the training set. Each scene-object pair contains randomized initial and target SE(3) poses, creating 30 out-of-distribution test cases. These configurations strictly avoid spatial overlaps with training data to validate cross-scene robustness.

\textbf{Evaluation of Grasp Pose Generation} As shown in Table 2 in the paper, for the ablation study on grasp generation, SGAP generates 10 grasp candidates per object pose using identical test conditions, resulting in 600 evaluation trials (30 cases × 10 attempts).

\subsection{Baseline Reproduction} 
\begin{itemize}
\item \textbf{LINGO}~\cite{jiang2024autonomous}: A scene-conditioned motion generator. As their open-source implementation lacks object trajectory synthesis, we reproduce the object interaction results using their paper's methodology. Specifically, it involves manually predefining the positions where the finger joints should grasp an object, and then the model generates motions to make the hand joints approach these predefined positions, thereby achieving a grasping effect. After the grasping frame, the object and the hand will maintain a fixed relative position, so as to realize the effect that the object moves along with the hand. Since it additionally requires predefined finger positions, we use the hand pose generated by our method as an additional condition. 

In the subsequent qualitative result demonstrations and the demo video, there are cases where the hand and the object are far apart. This is because the LINGO model has poor control ability over the motion space, and the hand movements fail to reach the predefined positions, which is also an inherent defect of LINGO. Since LINGO and TRUMANS are works from the same research group and use similar scene data, and LINGO mainly perceives local scenes (which are always similar), we did not perform additional fine-tuning of LINGO on TRUMANS.

\item \textbf{CHOIS}~\cite{li2024chois}: A object-aware motion model requiring pre-defined object paths. Essentially, CHOIS cannot perceive the scene. However, it can still accomplish the HOSI task as long as the trajectory of the object in the scene is provided in advance. We feed CHOIS with trajectories from Algorithm~\ref{alg: astar_path} to ensure comparable navigation inputs. Note for fair comparison, we fine-tune CHOIS on the TRUMANS dataset for adapting new objects.
\end{itemize}

\begin{figure*}[ht]
    \centering
    \includegraphics[width=\textwidth]{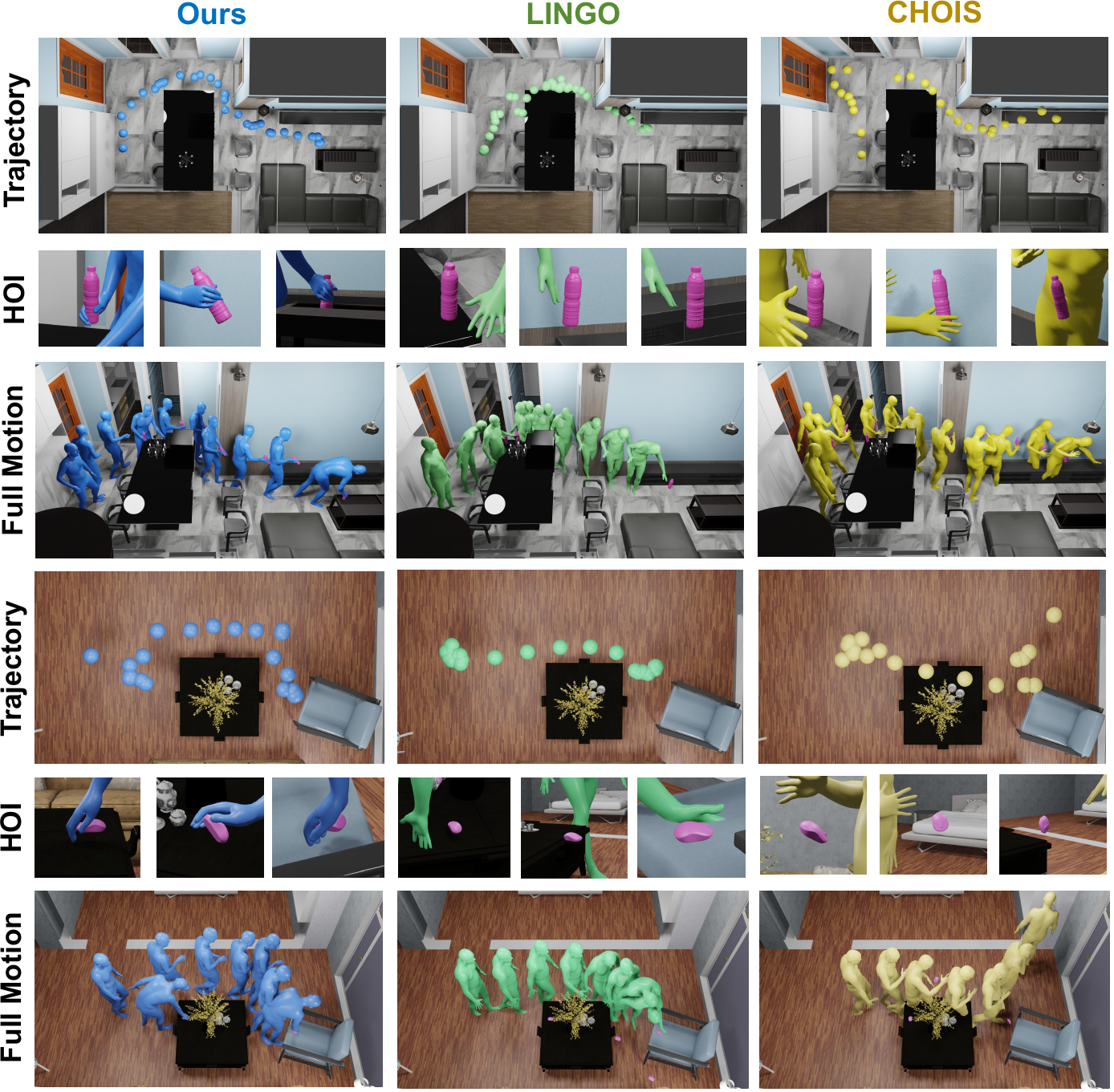}
    \caption{\textbf{Qualitative Results} In different scenes, three methods are required to generate the motion of picking up a bottle or a mouse from the same position and placing it at a target position. The “Trajectory” row is the generated human root joint trajectory. The “HOI” row shows the detailed hand-object contact at three different moments of picking up, transporting, and putting down. The “Full Motion” row is the complete motion display.}
    \label{fig: qualitative}
\end{figure*}

\begin{figure*}[ht]
    \centering
    \includegraphics[width=0.99\textwidth]{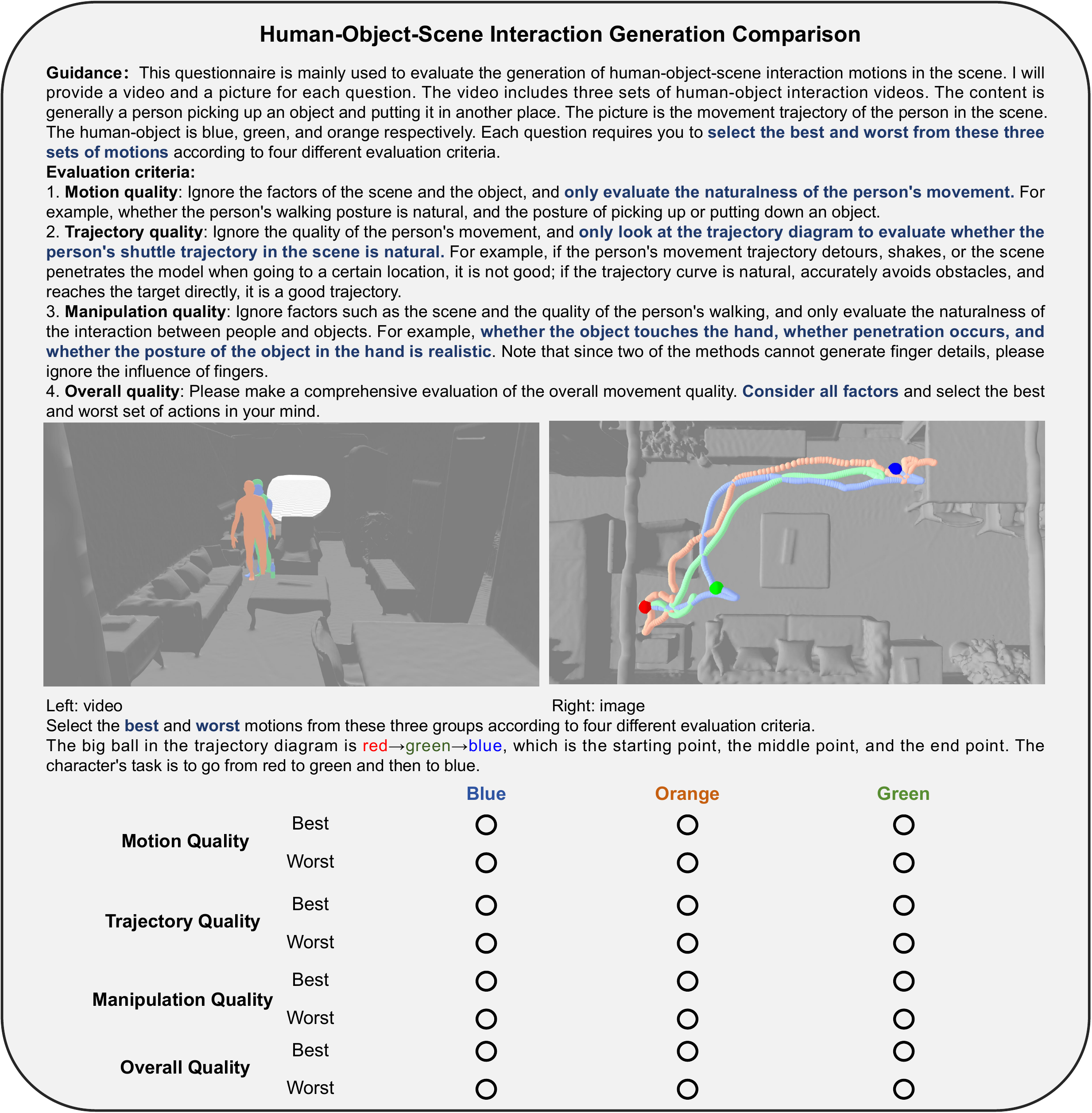}
    \caption{\textbf{Questionnaire Example.} When users participate in our user study, they will encounter a questionnaire similar to this one. Generally, they will be asked to answer three questions randomly selected from our question bank. We will calculate the score of each method based on their choices.}
    \label{fig: question}
\end{figure*}

\end{document}